\documentclass[runningheads]{llncs}
\usepackage[utf8]{inputenc}
\usepackage{graphicx}
\graphicspath{ {./images/} }
\usepackage{color}
\usepackage{graphicx}
\usepackage{booktabs}
\usepackage{longtable}
\usepackage{tabularx}
\usepackage{amssymb}
\usepackage{pifont}

\newcommand{\xmark}{\ding{55}}
\usepackage{rotating}
\usepackage{multirow}
\usepackage{caption}
\usepackage{subcaption}

\newcommand{\fastman}{{\fontfamily{lmtt}\selectfont FaSTM$\forall$N}}
\usepackage[
backend=biber,
style=numeric,
sorting=ynt
]{biblatex}
\PassOptionsToPackage{hyphens}{url}\usepackage{hyperref}
\usepackage{tikz}
\def\checkmark{\tikz\fill[scale=0.4](0,.35) -- (.25,0) -- (1,.7) -- (.25,.15) -- cycle;} 
\usepackage{wrapfig}
\usepackage{float}

\begin{document}
\title{Topology-Agnostic Detection of Temporal Money Laundering Flows in Billion-Scale Transactions}
\titlerunning{\fastman: Follow all Suspicious Trails of Money for all Nodes}
\author{Haseeb Tariq\(^{12}\)\orcidID{0000-0003-0756-3714}, Marwan Hassani\(^1\)\orcidID{0000-0002-4027-4351}}
\institute{Eindhoven University of Technology, Eindhoven, The Netherlands\\ \and
Transactie Monitoring Nederland, Amsterdam, The Netherlands \\
\email{m.h.tariq@tue.nl \quad m.hassani@tue.nl}}
\authorrunning{H. Tariq and M. Hassani}
\maketitle
\begin{abstract}

Money launderers exploit the weaknesses in detection systems by purposefully placing their ill-gotten money into multiple accounts, at different banks. That money is then layered and moved around among mule accounts to obscure the origin and the flow of transactions. Consequently, the money is integrated into the financial system without raising suspicion. Path finding algorithms that aim at tracking suspicious flows of money usually struggle with scale and complexity. Existing community detection techniques also fail to properly capture the time-dependent relationships. This is particularly evident when performing analytics over massive transaction graphs. We propose a framework (called \fastman), adapted for domain-specific constraints, to efficiently construct a temporal graph of sequential transactions. The framework includes a weighting method, using \(2^{nd}\) order graph representation, to quantify the significance of the edges. This method enables us to distribute complex queries on smaller and densely connected networks of flows. Finally, based on those queries, we can effectively identify networks of suspicious flows. We extensively evaluate the scalability and the effectiveness of our framework against two state-of-the-art solutions for detecting suspicious flows of transactions. For a dataset of over 1 Billion transactions from \textit{multiple} large European banks, the results show a clear superiority of our framework both in efficiency and usefulness.    

\end{abstract}

\section{Introduction}
\label{sec:introduction}

The United Nations Office on Drugs and Crime (UNODC) estimates that between 2 and 5\% of global GDP is laundered each year. That’s between EUR 715 billion and 1.87 trillion each year \cite{mloverview}. The human cost of this is even harder to estimate. Financial systems have not been able to effectively monitor and detect these money laundering networks. Consequently, some of the big banks have had huge fines imposed on them by the regulators. One of the major challenges banks face is that they only see their side of the money trail, whereas, a money launderer might be moving money via multiple banks. Most of the money laundering detection models have been developed keeping in mind single bank transaction data. When it comes to graph algorithms, this means that the banks are rarely interested in processing money trails with more than 2 hops \cite{dbj}, \cite{flowscope}.
\begin{figure}[h]
    \centering
    \includegraphics[width=0.8\textwidth]{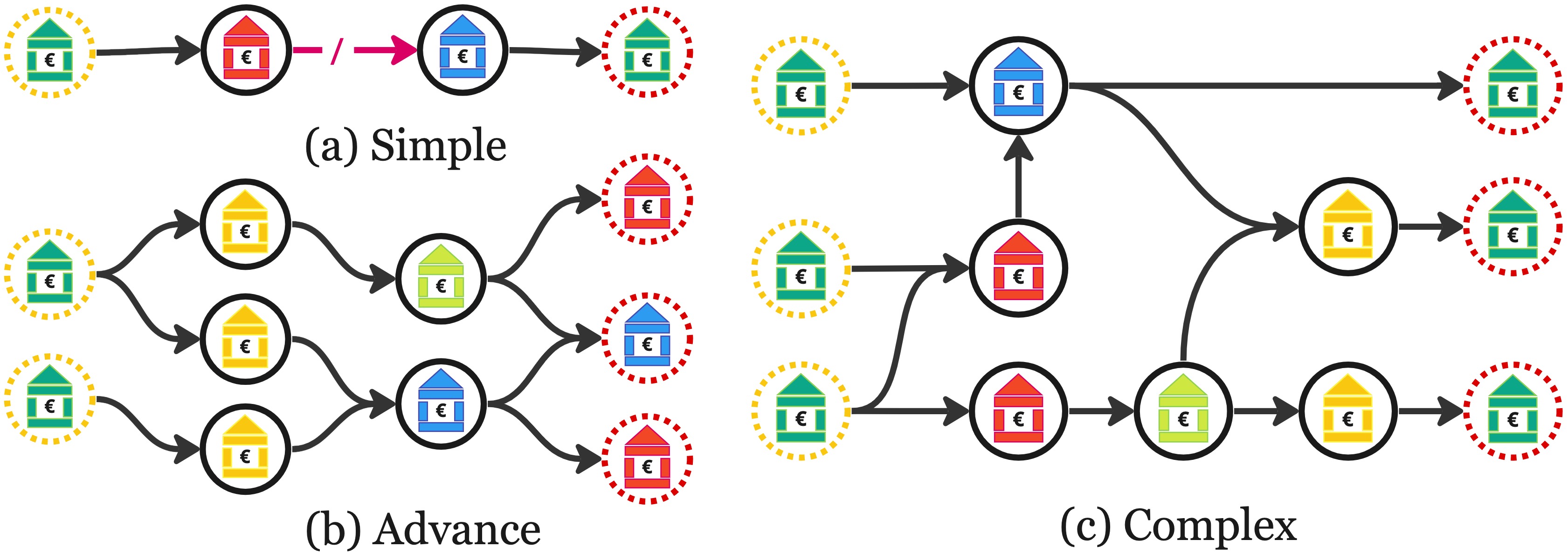}
    \caption{Examples of different levels of money trails. A circle represents a node or a bank account. The colors of the nodes represent the different banks the accounts are held in. The dotted orange outline on a node represents a dispense account; the dotted red outline a sink account. A directed arrow represents an edge or a transaction from a source to a destination node. The broken red arrow in (a) highlights that the trail is invisible to the green bank.}
    \label{fig:flow-complexities}
\end{figure}
The money trail shown in Fig. \ref{fig:flow-complexities}(a) is simple, as there is no account that sends money to multiple bank accounts, but still invisible to the green bank. The money starts flowing from the dispense account (the left most); and ends up in the sink account (the right most account), after making 2 hops using two other intermediate bank accounts. In the advance example in Fig. \ref{fig:flow-complexities}(b), there are multiple dispense; intermediate; and sink accounts. The shortest distance between every dispense and sink account is 3. For the complex flow in Fig. \ref{fig:flow-complexities}(c), the shortest distances between dispense and sink accounts are varying in lengths. The shortest paths are of lengths 2, 3, and 4.

Placement, Layering, and Integration \cite{mloverview} are the 3 commonly known phases of money laundering. Placement is when a criminal injects their ill-gotten money in the financial system, a bank for instance, without raising any suspicion. The criminal then starts to convolute the money trail by layering, for instance, by moving around the money among accomplices. Finally, in the integration phase, the criminal withdraws the money from seemingly legitimate bank accounts operating under a false pretext. The different ways a criminal can place, layer, and integrate money are called typologies. We will be using two similar sounding words \textit{typology} and \textit{topology} through out our paper. We have described typology here, by topology we always mean graph topology. There are a lot of \textit{known} typologies, with even countless variations. Smurfing (or structuring) \cite{smurfing} is one such typology, where a criminal deposits a large chunk of cash using (several) smaller deposits in multiple mule \cite{mule} accounts.

Criminals are coming up with new ways (or typologies) to trick the financial systems all the time. It is difficult for the Anti Money Laundering (AML) experts and modelers to keep up. Most of the modeling is therefore focused on capturing the \textit{known} unknown cases, and designed for single bank transaction data. Transaction Monitoring Netherlands (TMNL) \cite{tmnl} is an initiative where 5 major Dutch banks have joined forces. Together, they intend to tackle financial crime by collaboratively monitoring the banks' payment transactions for signs of money laundering and terrorism financing. Recently, there have been other similar initiatives \cite{acpr} \cite{cosmic} \cite{aurora} throughout the world. This creates the necessity of 1) dealing with longer (than 2 hops) money flows in a scalable way. And 2) capturing complex types of flows like the one shown in Fig. \ref{fig:flow-complexities}(c). Our framework, \textbf{F}ollow \textbf{a}ll \textbf{S}uspicious \textbf{T}rails of \textbf{M}oney \textit{for \textbf{a}ll} \textbf{N}odes ({\fastman}), deals with these limitations. With {\fastman} we are able to find the \textit{unknown} unknown cases by discovering new typologies. Our key contributions are:
\begin{itemize}
\item A scalable method to save and retrieve *all (see \( \Delta w \) in Definition \ref{def:temporalgraph}) possible money flows as a temporal graph. We show that adding the time dimension to the transaction graph could explode the space complexity. {\fastman} enables us to generate the temporal graph in a distributed manner; and to instantly jump to a point in time in the graph.
\item A novel weighting method that uses higher-order graph representation to quantify temporal relationships of sequential transactions. This helps us tremendously in removing insignificant edges from the temporal graph. Consequently, we are able to perform complex queries efficiently.
\item Topology agnostic detection of significant flows of money. {\fastman} is parameter-free when it comes to defining the topological properties like number of dispense, intermediate, and sink accounts; varying path lengths; and number of hops in a flow.
\item Evaluation on a large real-world dataset to show the effectiveness of our method in capturing an AML risk. High cash volume; transactions from or to high-risk countries (for instance, labelled by the Financial Action Task Force \cite{fatf}); unusual turnover volume for an industry; etc. are some of the examples of AML risks.
\end{itemize}

Section \ref{sec:related} discusses existing work. Sections \ref{sec:preliminaries} \& \ref{sec:method} introduce the definitions and {\fastman}. Section \ref{sec:experiments} details the experimental evaluation; and Section \ref{sec:conclusion} concludes the paper.

\section{Related Work}
\label{sec:related}
We are going to briefly review some of the previous research on the detection of money laundering networks using graph algorithms. We want to make a distinction here between money laundering and fraud detection methods. Although the two problems have some similarities, the focus of our research is strictly on detecting networks of money laundering. Specifically, we want to capture the manifestation of the anomalous behaviour of a money launderer in their transaction patterns. We are also not interested in detecting cycles \cite{10.14778/3229863.3229874} \cite{peng2022tdb} in our transaction graph. Money launderers use accounts in different banks and even different geographies, so that their deposits and withdrawals do not mimic a cyclic pattern. Mapping a UBO (or Ultimate Beneficial Owner) \cite{ubo} to the relevant bank accounts remains a big challenge for the banks \cite{ubo-data}. Finally, there are methods \cite{oddball} \cite{9762926} \cite{elliott2019anomaly} for detecting structural anomalies in transaction graphs. We believe that such methods are complementary to our framework and as such should not be compared against.

Graph modeling is a natural choice for transaction data. It is not trivial to make the more specific design decisions though. Decisions like, selection of node identifier (IBAN \cite{iban}, parent entity, or UBO \cite{ubo}); edge granularity (single or multiple transactions); edge weight and direction; or which attributes (if any) should be considered for nodes and edges. After finalizing the representation, the next task is to detect interesting communities. To deal with the possible scalability issues, \cite{SHARMA201744} and \cite{graph_partitioning} propose methods for parallel computing on massive graphs. \cite{study} presents a comparative study on some of the well-know community detection methods. The so called Louvain algorithm \cite{Blondel_2008} is one of the most popular modularity-based \cite{modularity} community detection methods. Recently a more advanced version, called the Leiden algorithm, has been proposed in \cite{leiden} which we use in our framework. In \cite{wagenseller2017size}, several algorithms are evaluated by taking into account the community size distributions. They show that the Dunbar's number \cite{dunbar1993} of 150 is relevant in contexts other than social sciences. The community size is even more important in the context of AML modeling. Anomalous communities have to be analysed by the AML experts eventually. A big community size, in some cases, makes it humanly impossible to analyze. 

A comprehensive survey has been published in \cite{akoglu2014graphbased} for anomaly detection in data represented as graph. By design, most community detection methods would detect communities with dense intra-connections and light inter-connections. For a transaction dataset, this would result in communities with short paths. In most cases, paths of 1 or 2 hops are observed. This makes the simpler methods like \cite{oddball} or \cite{ego2} more suitable for detecting anomalous communities with close neighbors. Due to the strict scrutiny of the banks by the data and financial regulators lately, the AML models have to be void of any kind of biases. This limits the number of data points that can be used in AML models. We are talking of, for instance, geographic and demographic biases. Guilt by association (or association fallacy) is one such risk that the AML modelers have to be aware of. A recent paper \cite{supervised} proposes a method that can fall under this trap.

Detecting (dense) flows of money is different from detecting communities in a transaction graph. A flow, by definition, has a temporal order. In the context of AML modeling a \textit{dependent} flow of transactions is, in most situations, more interesting then the immediate interactions a bank account has. For motifs in temporal graphs, \cite{Paranjape_2017} introduces efficient algorithms. Motif queries are bounded and deterministic, whereas our goal is to capture networks with any possible composition. We will now introduce the 2 recent methods we are going to evaluate {\fastman} against. The first one is called FlowScope \cite{flowscope}; and the second one \cite{dbj} we are going to call the Database Joins method or DBJ from this point on. A major shortcoming of FlowScope is that its design is strictly relevant for a single bank dataset (see Section \ref{experimentsetting}). Table \ref{table:methods-comparison} summarizes the key differences for the 3 methods. In the table, by dynamic grouping we mean that with DBJ you have to explicitly define how you want to group the detected paths as flows. The definition is determined by the type 1 and type 2 motif types they describe in their paper.

In a recent paper \cite{geometry}, the authors summarize 5 \textit{motivations} or topological rules to detect behaviours of money laundering. We are going to summarize the relevant ones here:
\begin{itemize}
\item \textit{Motivation 1}: Few (but repeated) interactions with the accomplices,
\item \textit{Motivation 2}: Complicating the money trail by making several hops,
\item \textit{Motivation 3}: Convoluting it further by having more interactions among the accomplices, and,
\item \textit{Motivation 4}: Breaking down big transactions into many small transactions.
\end{itemize}
With {\fastman} we address these motivations by making as few assumptions as possible. We argue that the designs for FlowScope and DBJ, do not take into account \textit{all of these} motivations.
\newcolumntype{L}{>{\centering\arraybackslash}m{2cm}}
\begin{table}[h!]
\centering
\begin{tabular}{|| c | L | L | L | L ||} 
 \hline
  & Dynamic Grouping & Parameter-free for \# of hops & Complex Flows & Suitable for multi-bank data \\
 \hline\hline
    \textbf{DBJ}~\cite{dbj} & \xmark & \xmark & \xmark & \checkmark \\
    \textbf{FlowScope}~\cite{flowscope}  & \checkmark & \xmark & \xmark & \xmark \\
    \textbf{{\fastman}} (Ours) & \checkmark & \checkmark & \checkmark & \checkmark \\
 \hline
\end{tabular}
\caption{Features Comparison of State-of-the-art AML approaches and {\fastman}}
\label{table:methods-comparison}
\end{table}

\section{Preliminaries}
\label{sec:preliminaries}
In Section \ref{sec:related}, we mentioned solutions for detecting communities of entities; and for detecting flows of money. Detecting \textit{communities of flows} is, to the best of our knowledge, a novel approach that we are proposing in this paper. {\fastman} is able to detect closely related and interdependent sequences of transactions; which are essentially communities of flows. In this section we will describe some of the related concepts.
\subsection{Temporal and Higher Order Graphs}
\label{sec:temporal-second}
\begin{wrapfigure}[8]{r}{0.5\textwidth}
    \centering
    \includegraphics[width=0.5\textwidth]{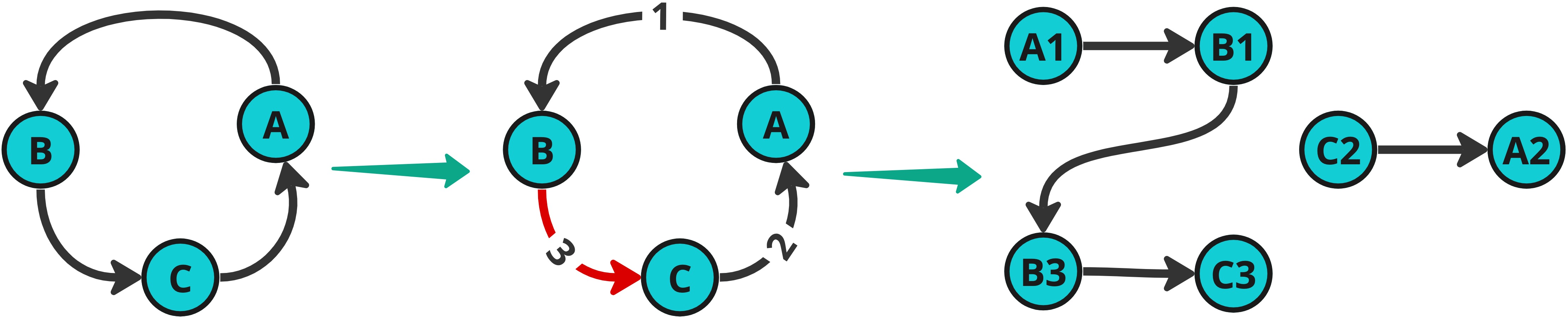}
    \caption{Temporal path traversal}
    \label{fig:temporal}
\end{wrapfigure}

Temporal information can be incorporated into data in different ways. The dynamic graphs mentioned in \cite{akoglu2014graphbased} are (mostly) series of static snapshots of graphs, at different moments in time. Time can also be modeled as layers in a multiplex graph \cite{multiplex}. There are several data modeling techniques for time-dependent graphs, surveyed in \cite{temporal}. Different types of temporal graphs have different applications. We are mostly interested in temporal graphs where the paths can only be traversed in a chronological order. In most money laundering typologies, the money is only pushed forward to the next destinations once it has been received from the previous parties. Therefore, for {\fastman}, we build a static time-dependent graph by creating copies of nodes \cite{10.14778/2732939.2732945} \cite{spanning} \cite{temporal-shortest}. Specifically we want to invalidate the path shown in the left most graph of Fig. \ref{fig:temporal}. The flow looks interesting because of the cyclic behaviour. On a closer look (middle graph), after taking into account the chronological order of transactions, the cyclic behaviour is not there anymore. The aim is to convert the left most graph to the right most graph, by respecting the temporal order. It can be observed that the number of nodes and edges could potentially explode with this type of representation. We show in Section \ref{sec:method} how we limit this explosion by taking key AML knowledge into account.

\begin{wrapfigure}[8]{r}{0.5\textwidth}
    \centering
    \includegraphics[width=0.5\textwidth]{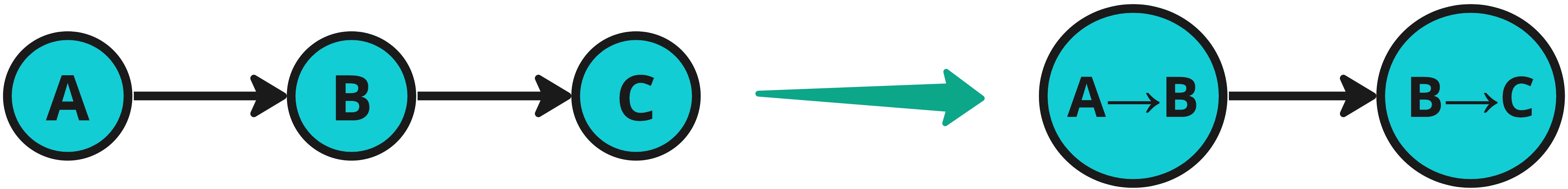}
    \caption{From first to second order representation}
    \label{fig:second}
\end{wrapfigure}

Our method for calculating the edge weights is based on higher order graph representations, and the principles explained in \cite{Scholtes_2017}. An example of the \(2^{nd}\) order representation can be seen in Fig. \ref{fig:second}. On the left hand side we have the \(1^{st}\) order representation, where account A sends money to account B, which then transfers it to C. On the right hand side we have the \(2^{nd}\) order representation, a transaction from A to B is followed by a transaction from B to C. With this representation we want to incorporate a (short) memory footprint of the sequential transactions. Consequently, the more dominant (or frequent) a memory is; the stronger the interdependence of the sequential transactions becomes. For our experiments, we did not consider higher (than \(2^{nd}\)) order (\(3^{rd}\), \(4^{th}\), ..., \(n^{th}\)); or multi-order representations. We believe that evaluating against different representations is in itself a big enough effort for an exclusive research topic. We plan to cover that topic in our future research.

\subsection{Dataset}
\label{sec:dataset}
Our dataset includes around \textbf{1.1 billion} transactions from 5 large Dutch banks. We can not reveal (see \textit{reasons} in Section \ref{sec:experiments}) much about the data because of its sensitive nature. The input data looks like the example included with the shared code (mentioned in Section \ref{sec:experiments}). In that dataset, \textit{id} represents the unique identifier for a transaction; \textit{transaction\_timestamp} the (Unix) time \cite{unixts} for the transaction; and \textit{source} and \textit{target} are the account identifiers for the sender and the receiver of the \textit{amount}, respectively. These are the bare minimum data points required to run {\fastman}. We will be using the dataset from the shared code as a running example to explain the different parts of {\fastman}.

\section{The Method}
\label{sec:method}
\begin{figure}[h]
    \centering
    \includegraphics[width=1\textwidth]{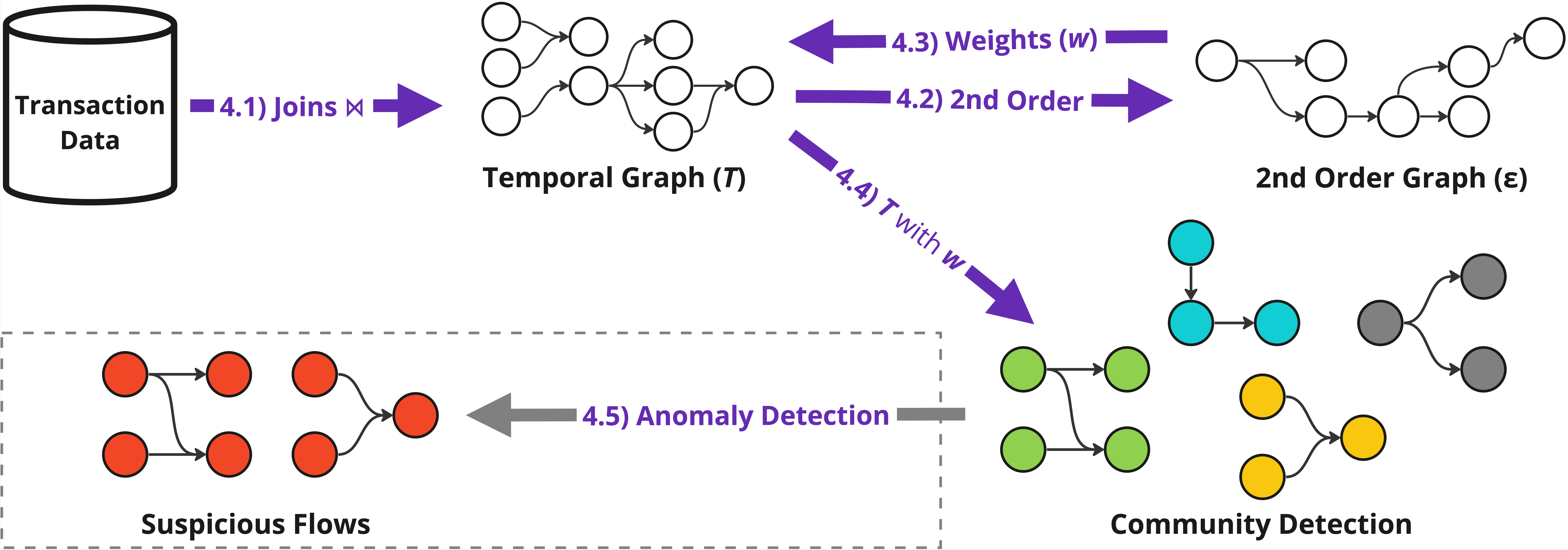}
    \caption{End-to-end architecture diagram of the {\fastman} framework. The grey dashed box is not an \textit{integral} part of the framework.}
    \label{fig:framework}
\end{figure}

The architecture diagram of {\fastman} is explained in Fig \ref{fig:framework} with pointers to the sections where the relevant steps are introduced. The entire framework is implemented in Spark \cite{spark} and its Graphframes \cite{graphframes} engine, which make the different computations in {\fastman} highly distribute-able and scalable. 

\subsection{Temporal Graph Creation}
We create the temporal graph of sequential transactions by making inner joins on our dataset. The transactions are processed iteratively, one day at a time. This ensures that 1) the memory footprint of the inner joins is minimized; and 2) the output graph can be saved and retrieved (based on time queries) from the distributed file system \cite{5496972} instantly. Here is an example of sequential transactions: account A sends money to account B; account B then, \textbf{afterwards}, sends some money to account C. The intuition here is that for some cases, account B might have only sent (a part of the) money to account C because it had earlier received (a part of) it from account A. Naturally, this will not be true for many other cases. In Section \ref{sec:weight}, we are going to remove all such insignificant relationships.
\begin{definition} [Temporal Graph of Sequential Transactions]
\label{def:temporalgraph}
A static, attributed, and directed graph \( \mathcal{T} = \{V, E, X\} \) comprises 1) a node set \( V = \{v_i\}_1^n \), where a node is represented using a unique transaction identifier \( i: 1...n \), with \( n \) as the number of transactions. 2) An edge set \( E = \{e_{s,d}\} \) where an edge going from node \( v_s \) to \( v_d \) is denoted as \( e_{s,d} = (v_s, v_d) \), with \( v_s \) representing the \textbf{source} and \( v_d \), the \textbf{destination} node. 3) The nodes attribute matrix is represented as \( X = [\textbf{x}_i]_n\); where vector \( x_i = [f_i, b_i, t_i, a_i] \). For vector \( x_i \), \( f_i \) is the \textbf{from} account that sends the amount; \( b_i \) is the \textbf{beneficiary} account that receives the amount;  \( t_i \) is the \textbf{timestamp} of the transaction; and \( a_i \) is the transaction \textbf{amount}. For \( e_{s,d} \) the corresponding vectors for source and destination attributes are represented as \( \textbf{x}_s \) and \( \textbf{x}_d \), respectively. There exists an edge \( e_{s,d} \) \( iff \):
    i) \( b_s = f_d \), and
    ii) \( t_s < t_d < (t_s + \Delta w) \),
 where \( \Delta w \) is the time window parameter for "looking ahead" in time.
\end{definition}

\begin{wrapfigure}[10]{r}{0.5\textwidth}
    \centering
    \includegraphics[width=0.5\textwidth]{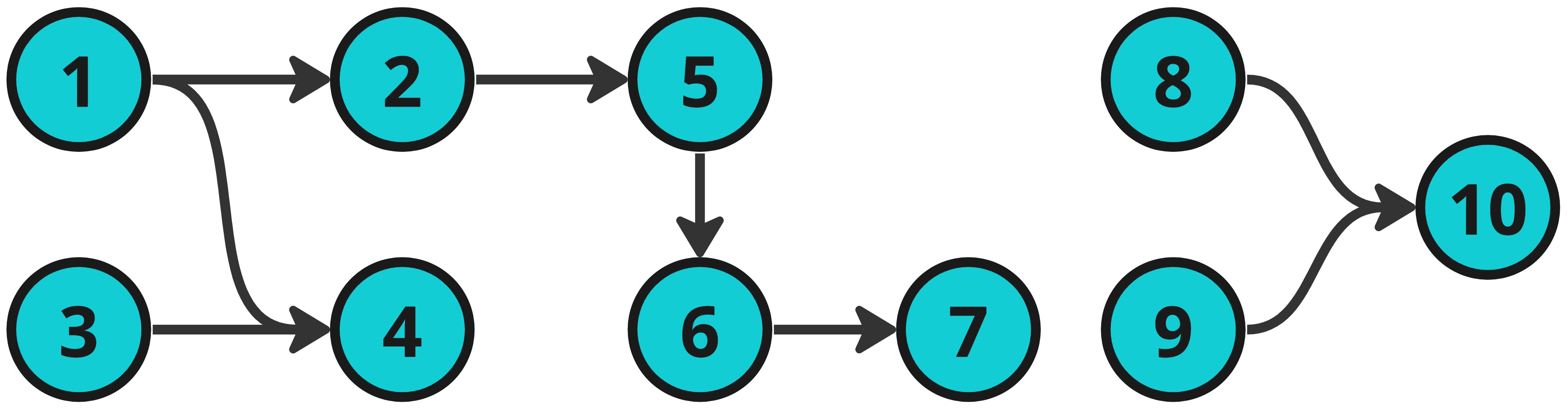}
    \caption{Temporal graph for the example dataset}
    \label{fig:temporal-example}
\end{wrapfigure}
The file partitioning is based on source and destination nodes' dates, which makes it possible to time travel to any moment in the graph instantly. The details of this implementation can be found in the shared code mentioned in Section \ref{sec:experiments}.

This step can produce a huge number of edges in \( \mathcal{T} \). We limit this by the \( \Delta w \) parameter. There is a cost associated with holding the money in an account and not passing it forward. Therefore while investigating a money trail, for most typologies, any trail would become uninteresting if the time interval between consecutive transactions is longer than \( \Delta w \) days. We use a very conservative \textit{undisclosed} (see \textit{Reason 1} in Section \ref{sec:experiments}) value for \( \Delta w \), therefore, we end up with around \textbf{25 billion} edges. Fig. \ref{fig:temporal-example} shows the resulting temporal graph for our running example.

\subsection{\(2^{nd}\) Order Graph Creation}
\begin{wrapfigure}[11]{r}{0.5\textwidth}
    \centering
    \includegraphics[width=0.5\textwidth]{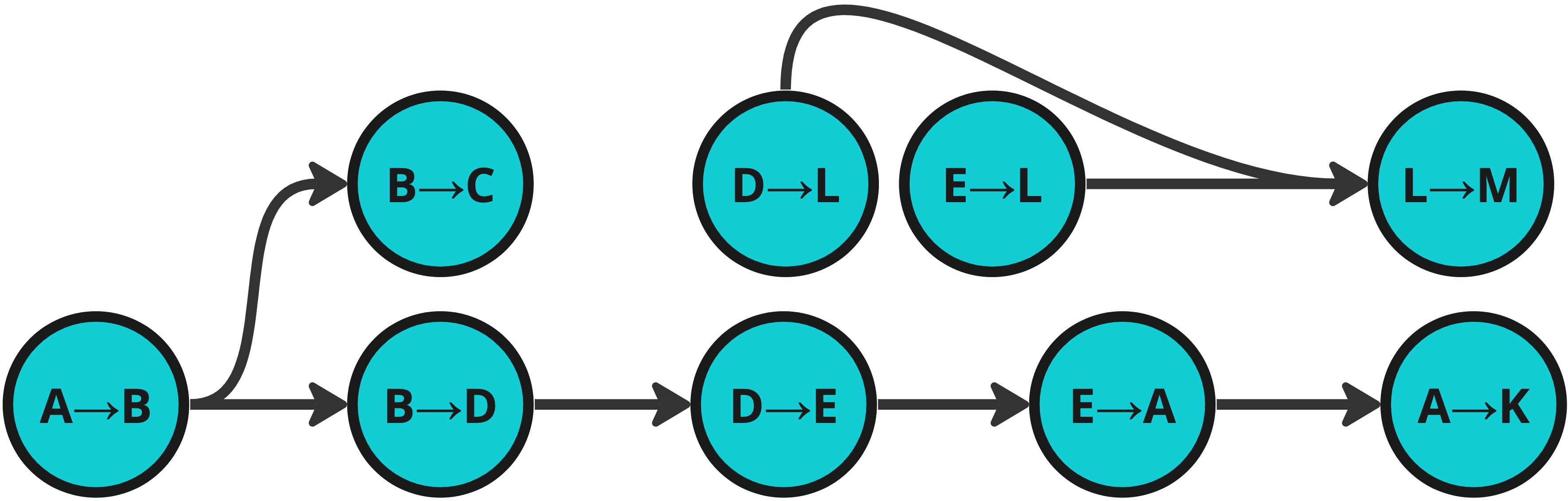}
    \caption{\( 2^{nd} \) order transformation for the example dataset}
    \label{fig:second-example}
\end{wrapfigure}
The \(2^{nd}\) order graph \( \mathcal{S} \) is \textbf{derived} from the already generated temporal graph \( \mathcal{T} \). Instead of using the unique transaction index as the node identifier; the nodes in this graph will be represented by the concatenation of \( (f_i, \rightarrow, b_i) \). The idea behind this step is that with \( \mathcal{T} \), we have included all the sequential transactions within \( \Delta w \).
In reality, most of those sequential transactions will not be dependent on each other; or some will be more dependant than others. Using \( \mathcal{S} \), we want to quantity this relationship for later stages. The transformation from \( \mathcal{T} \) to \( \mathcal{S} \) for the example dataset can be seen in Fig \ref{fig:second-example}.

\subsection{Weight Calculation}
\label{sec:weight}
In this step we quantify the strength of the \( 2^{nd} \) order relationships, that we will later apply on \( \mathcal{T} \). The formal definition is as follows.
\begin{definition} [Co-occurrence Weight]
\label{def:weight}
Using \( \mathcal{S} \), the co-occurrence weight between a source node \( A \rightarrow B \) and a destination node \( B \rightarrow C \) is calculated as, \[ \mathcal{W}(A \rightarrow B, B \rightarrow C) = max(\mathcal{P}(A \rightarrow B, B \rightarrow C), \mathcal{P}^{\prime}(A \rightarrow B, B \rightarrow C)) \] where,
\[ \mathcal{P}(A \rightarrow B, B \rightarrow C) = \frac{|\mathcal{S}(A \rightarrow B\sim B \rightarrow C)|}{|\mathcal{S}(A \rightarrow B\sim B \rightarrow [*])|} \] and,
\[ \mathcal{P}^{\prime}(A \rightarrow B, B \rightarrow C) = \frac{|\mathcal{S}(A \rightarrow B\sim B \rightarrow C)|}{|\mathcal{S}([*] \rightarrow B\sim B \rightarrow C)|} \] where, \( [*] \) represents \textbf{any} account and \( \sim \) represents directed adjacency from the left to the right node(s).
\end{definition}

\(\mathcal{P}\) is the weight calculated from the source perspective; and \(\mathcal{P}^{\prime} \) from the destination perspective. We will have the maximum value of 1 for \(\mathcal{P}\), in the following scenario: every time A sends money to B; B moves (at least some of) the money \textit{exclusively} to C. For \(\mathcal{P}^{\prime} \), we will have the maximum value of 1 in the following scenario: every time C receives money from B; B had earlier received (at least some of) the money \textit{exclusively} from A. It can be noticed here that again this is a very conservative calculation. By taking the maximum of \(\mathcal{P}\) and \(\mathcal{P}^{\prime} \), we are giving the edge every chance to have a decent weight. The weights are calculated on the entire available dataset. The only time an edge will have a low weight is when extremely infrequent co-recurring behavior is observed for (the accounts involved in) the sequence of transactions. This is inline with \textit{Motivation 1} (Section \ref{sec:related}): in case of few \textit{direct} interactions, the \( 2^{nd} \) order interactions would still have a relatively high weight. The weights can now be applied to the original temporal graph  \( \mathcal{T} \) as follows:
\[ \mathcal{W}(e_s, e_d) =  max(\mathcal{P}(f_s \rightarrow b_s, f_d \rightarrow b_d), \mathcal{P}^{\prime}(f_s \rightarrow b_s, f_d \rightarrow b_d)) \]
After applying the weights on the edges of \( \mathcal{T} \), we filter out all the edges with a very low \textit{undisclosed} (see \textit{Reason 1} in Section \ref{sec:experiments}) value. This helps us in reducing the search space (see Table \ref{table:execution-times}), by removing superficially connected transactions.
\subsection{Community Detection}
\begin{wrapfigure}[5]{r}{0.5\textwidth}
    \centering
    \includegraphics[width=0.5\textwidth]{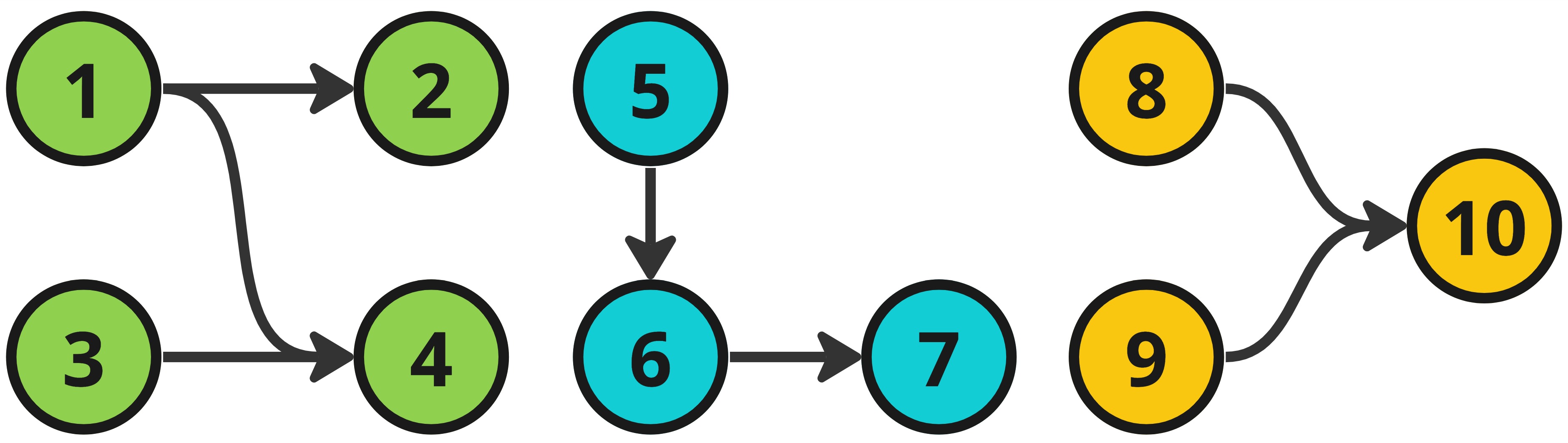}
    \caption{Communities detected for the example dataset}
    \label{fig:communities}
\end{wrapfigure}
In this step we pass the weighted and directed edges of \( \mathcal{T} \) to the Leiden algorithm \cite{leiden}. In addition to guaranteeing better quality communities, the algorithm also takes into account the directions and weights of the edges. For the example dataset, the algorithm identifies the communities shown in Fig. \ref{fig:communities}. Using the \(max\_comm\_size \) (maximum community size) parameter, provided with their library, we can also limit the number of transactions we want to include in a single flow (or community). This is the final step of {\fastman}, which produces communities of significant and interdependent money flows.

\subsection{Anomalous Communities Detection}
\label{sec:glad}
Strictly speaking, this is not a part of the framework. Once we have the communities of significant and dependent money flows from the last step, we can identify the suspicious networks by either applying 1) traditional anomaly detection methods \cite{10.1145/1541880.1541882} on the (summarized) vector representations of the money flows; or 2) any of the deep learning based graph level anomaly detection (or GLAD) \cite{glad} methods. For our experimental evaluation we flag the communities of money flows as suspicious, based on the following AML risk criteria:
\begin{itemize}
\item \textit{C1}: The dispense and sink accounts should belong to certain (\textit{undisclosed}) segments, \textit{and},
\item \textit{C2}: The percentage of dispensed amount that has been sunk should be within a certain (\textit{undisclosed}) range, \textit{and},
\item \textit{C3}: The maximum flow, respecting the temporal order, of money should be greater than a certain (\textit{undisclosed}) threshold
\end{itemize}

The rationale behind concealing certain details about the criteria is explained in Section \ref{sec:experiments}. 

\section{Experimental Evaluation}
\label{sec:experiments}

In this section we will show the superiority of {\fastman} over FlowScope~\cite{flowscope} and DBJ~\cite{dbj}, in terms of scalability and effectiveness. The implementation for the entire framework, using an example transaction dataset is shared here: \url{https://github.com/mhaseebtariq/fastman/blob/main/fastman.ipynb} The part that has not been shared is the AML risk calculation (Section \ref{sec:glad}). We will also conceal certain details about the detected flows. The reasons behind these decisions are as follows:
\begin{itemize}
\item \textit{Reason 1}: By learning about the details of a specific risk calculation, criminals can try to work around it to avoid detection.
\item \textit{Reason 2}: Some details in the detected flow, for instance transaction amounts and name of the banks, can alert a criminal to be on the run.
\end{itemize}

\subsection{Experimental Setting}
\label{experimentsetting}
 With graph \( \mathcal{T} \) we can detect all the 2-hop flows detected by DBJ. We can actually extend DBJ and traverse many more hops than just 2. This is because our architecture is scalable and benefits from the distributed computing engine of Graphframes. This makes the \textit{temporal graph representation} of {\fastman}, a significantly improved version of DBJ. We will evaluate {\fastman} against this already more scalable version of DBJ. We will refer to this version as DBJ* in the rest of the paper. We were unable to obtain the dataset used for generating the results in DBJ, even after contacting the authors. This is partially understood given the aforementioned \textit{Reason 2}. The motif type we have chosen for evaluating DBJ is the \( 2^{nd} \) one, i.e. 1 dispense account sending money to multiple [*] intermediate accounts; which then sink the money to 1 bank account. The dispense and sink accounts should belong the same segments as in the risk criterion \textit{C1}. The type 2 motif for 3 hops can be configured as \textbf{1} $\rightarrow$ [*] $\rightarrow$ [*] $\rightarrow$ \textbf{1}, and so on. We did not choose type 1 as, for hops greater than 2, the motif could be configured in numerous ways. You can not only restrict an account in different layers to 1; but can also restrict multiple accounts at multiple locations to 1.

The $k$-partite graphs mentioned in FlowScope have to be constructed with the perspective of the middle accounts, referred to in the paper as \( \mathcal{M} \). The middle accounts belong to the \textit{pass-through} bank for which the model is being developed. Extending their design to multiple configurations of \( \mathcal{A} \) (source money laundering accounts), \( \mathcal{M} \), and \( \mathcal{C} \) (right destination accounts) would require substantial efforts. Especially from the perspective of scalability. For instance, if you have data from 5 banks, you can have 31 different combinations of banks you can assign as the pass-through bank(s). Ignoring this design flaw in FlowScope, we are going to focus on its inability to detect the \textbf{complex} money flows introduced in Fig. \ref{fig:flow-complexities}. This is also explicitly built into its design, which does not require any proof. We will assume that it can produce all the flows detected by {\fastman}, after running it multiple times with different values of \textit{k}. By showing some \textbf{real} examples, we will demonstrate that after applying the risk score on those flows, some (parts) of the flows will go undetected (see Fig. \ref{fig:cases}).
 
\subsubsection{Pre-processing}
For our experiments, we create the baseline input data by filtering out some transactions. We do it in the most conservative manner. We calculate the connectedness for each of the nodes (or bank accounts); and then remove the bank accounts from the transaction data that are \textit{extremely} highly-connected. We only do that for the top 0.0001\% accounts. This results in a reduction of the actual dataset of 1.1 billion transactions, to around 510 million transactions. In contrast, DBJ makes several pre-processing steps resulting in a huge size reduction. They merely keep around \textbf{0.002\%} of their original transactions. By doing so, even for a very specific typology like smurfing (see \textit{Motivation 4} in Section \ref{sec:related}), a lot of interesting flows can go undetected by DBJ. Even when covering 100\% of the smurfing cases, such extensive filtering makes DBJ infeasible for detecting other AML typologies. 
\subsubsection{Functional Evaluation}
The flows are going to be detected for the data in a specific time window. The selection of the suspicious flows is based on the \textit{risk criteria} described in Section \ref{sec:glad}. The exact same risk criteria are considered for all competing methods. There are numerous ways the AML modelers flag a case, using countless criteria, as \textit{true positive}. There is a huge cost associated with investigating a single case. In our experiments, flows from {\fastman} and DBJ* will be flagged as suspicious if they fulfill our \textit{risk criteria}. The criteria are so specific that they make every qualifying flow worth investigating, no matter which method was used to detect that flow. We are therefore going to focus on the coverage and novelty of the suspicious flows, in terms of topological diversity. We will show that with a method like DBJ, because of its naïve grouping of the flows 1) an AML investigator will have to go through a lot of cases with overlapping bank accounts, and 2) those cases would still not include all the accounts that should have been flagged.

\subsection{Results}
For all the experiments and for all competing methods, we have used a Spark cluster with the same hardware and software specifications. We will go through the significant (comparative) results in this section.
\vspace{-0.2cm}
\subsubsection{{\fastman} Execution Times}
Table \ref{table:execution-times} shows how different stages of {\fastman} explodes and then implodes the search space.
\begin{table}[h!]
\centering
\begin{tabular}{||c | c | c ||} 
 \hline
 \textbf{Step} & \textbf{Transactions} & \textbf{\( \mathcal{T} \) Edges} \\
 \hline\hline
    Initial state & 1.1 billion & -  \\
    Pre-processing & 510 million & - \\
    \( \mathcal{T} \) creation & 475 million & 25 billion \\
    Remove weak edges & 325 million & 2.3 billion \\
 \hline
\end{tabular}
\caption{Space explosion and implosion after each step}
\label{table:execution-times}
\end{table}
In terms of the number of transactions, we merely doubled that quantity for the (final) number of edges in \( \mathcal{T} \). Throughout the framework we are mostly performing join and aggregate operations in a distributed manner. The most complex operation is the Leiden algorithm \cite{leiden}. For most applications, the algorithm is said to be (empirically) close to linear time with respect to the number of edges. We can distribute this computation by running the algorithm in parallel for each window.
\begin{figure}[h]
    \centering
    \includegraphics[width=0.6\textwidth]{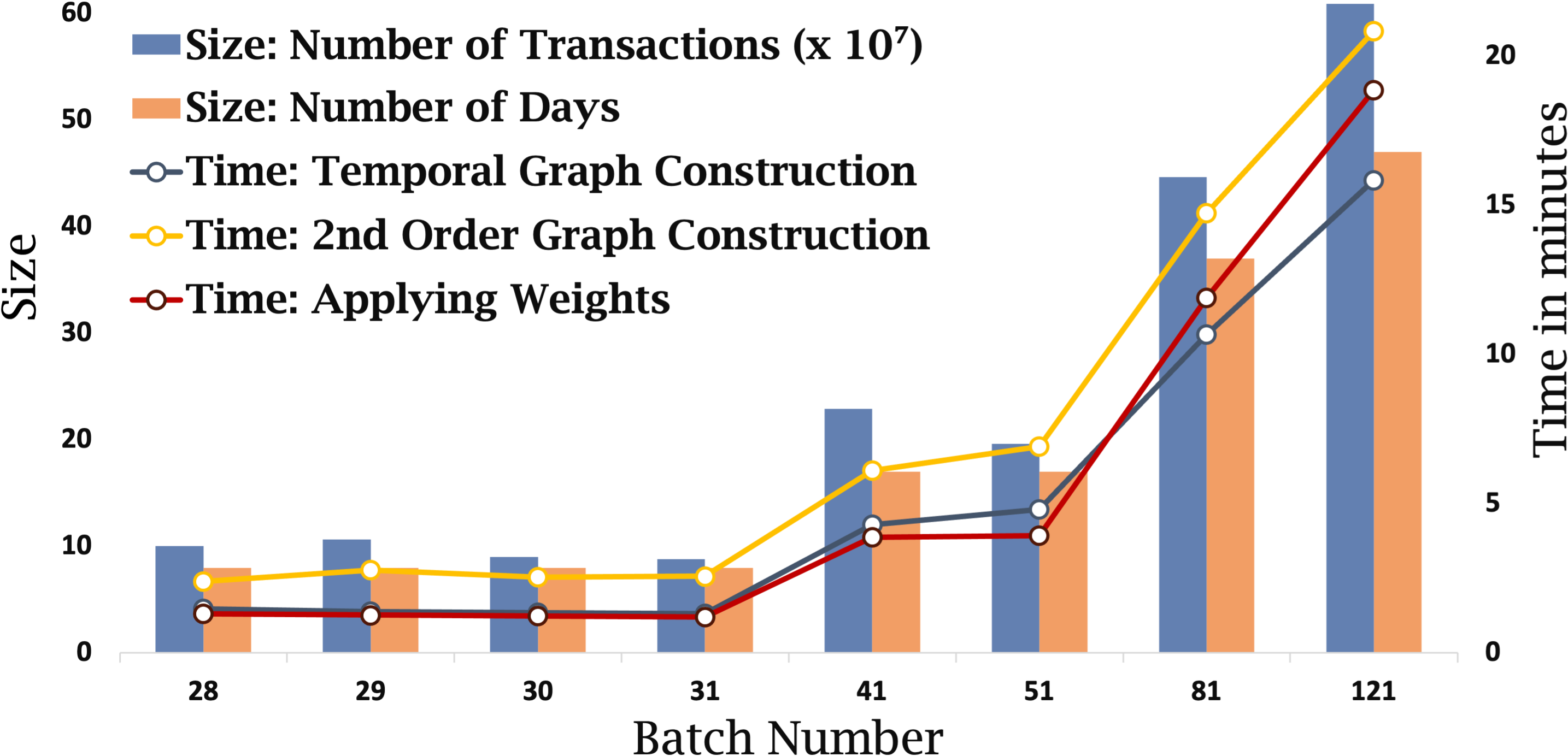}
    \caption{Runtimes for batches with different number of days in the data}
    \label{fig:runtimes}
\end{figure}

All the steps in {\fastman} except the last one, for (risk) scoring and picking the suspicious communities, are model and typology agnostic. This means that one can run those steps once and use the output to build many flavours of models, covering several risks. We ran an experiment for reporting the runtimes for those steps. We did it in such a way that every batch of new transactions was processed and appended to the already processed output from the previous batches. It can be seen in Fig. \ref{fig:runtimes} that the runtime of each step is (almost) linear with respect to the number of transactions in a batch. Over time, processing more transactions (or number of days of transactions) also does not seem to have a compounding effect on the total runtime.

With this level of scalability, {\fastman} can be seen as a back-end for building AML models. For time complexity, we are therefore, only going to compare the runtimes for the community detection part of {\fastman} versus the motif searching part of DBJ*. We are then going to compare the results by applying the exact risk scoring on the outputs of both.
\vspace{-0.3cm}
\subsubsection{Flow Detection}
\begin{figure}[h]
    \centering
    \includegraphics[width=0.6\textwidth]{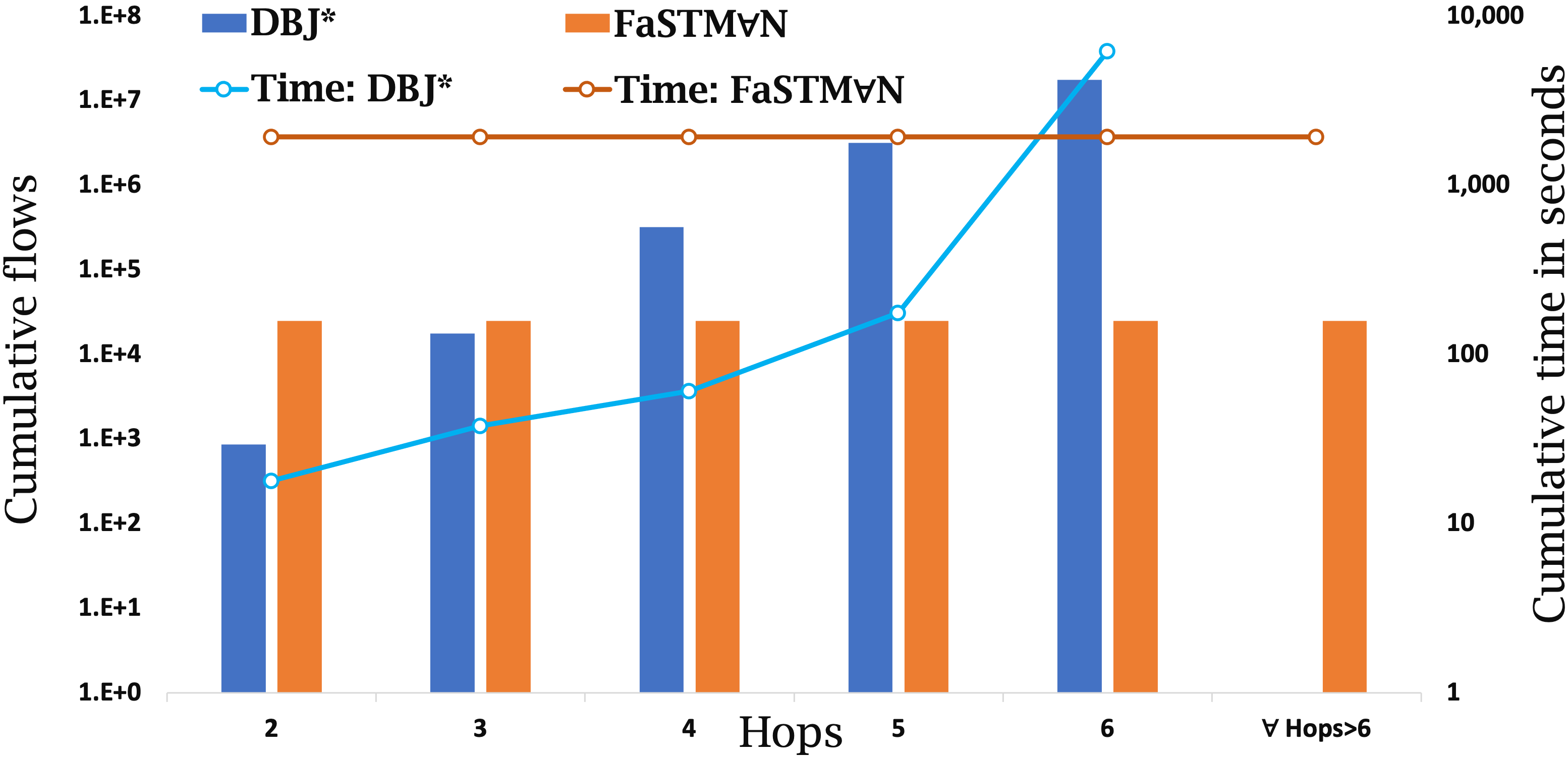}
    \caption{Number of detected flows for DBJ* exploding with every hop. Both y-axes scales are logarithmic.}
    \label{fig:dbj-complexity}
\end{figure}

Fig. \ref{fig:dbj-complexity} shows that the number of flows for DBJ* explode as we traverse more hops with the motif queries, resulting in (most) flows with overlapping transactions. This is not the case with {\fastman}, as the Leiden algorithm detects disjoint communities. 
We also only have to detect the communities, with any number of hops, once. After hop 5, the search space starts to become exponentially large for DBJ*. This has implications on the runtime, as can be seen in Fig. \ref{fig:dbj-complexity}.
\vspace{-0.2cm}
\subsubsection{Suspicious Flow Detection}
Flagging the detected flows as \textit{suspicious}, after applying the \textit{risk criteria}, involves expensive computations. Having fewer flows to flag is therefore ideal for efficiency.
\begin{figure}[h]
    \centering
    \includegraphics[width=1\textwidth]{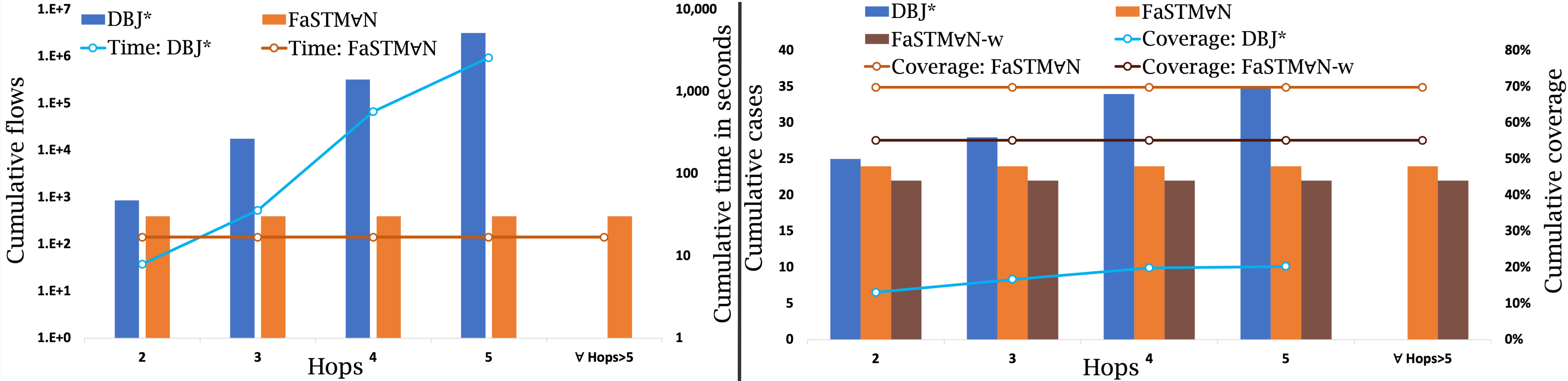}
    \caption{Suspicious flow detection comparisons. (a) A comparison for runtimes. Both y-axes scales are logarithmic. (b) A functional comparison for the suspicious flows. Higher coverage with lower number of cases is the desired outcome.}
    \label{fig:functional}
\end{figure}
Fig. \ref{fig:functional}(a) shows how with more hops, the runtime of DBJ* for flagging the flows increases. For {\fastman} it is very low, as 1) we only have to flag the flows which contain the segments described in criterion \textit{C1}. And 2) {\fastman} generates (substantially) fewer flows to begin with, as shown in Fig. \ref{fig:dbj-complexity}. Fig. \ref{fig:functional}(b) shows the coverage achieved by the different methods. In the experiments we also have a version called {\fastman}-w, where we did not include the weights in the community detection step. Coverage of a method is defined as the percentage of suspicious bank accounts found by that method, out of the total number of suspicious accounts found by all three methods. The results clearly show that by including weights, {\fastman} is able to improve  case coverage from 55\% to 70\% with nearly the same number of cases.

The trends shown in Fig. \ref{fig:functional} clearly indicate that, for DBJ* to achieve a coverage score close to {\fastman}, it would have to be run with numerous different motif configurations. This would result in astronomically high number of flows. Consequently, flagging those flows would take unrealistic time. Even if achievable, the number of cases would make it practically impossible to investigate. This is because most of the flows will have repeated accounts and transactions.

\begin{figure}[h]
    \centering
    \includegraphics[width=1\textwidth]{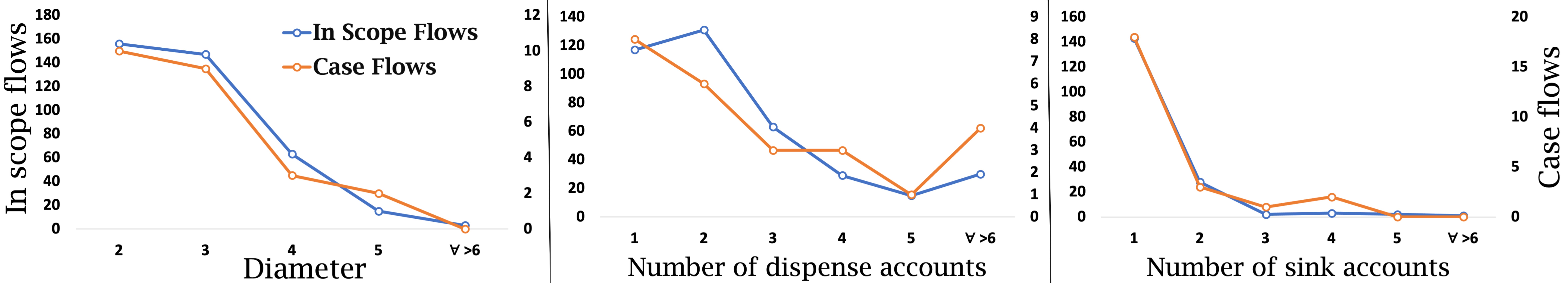}
    \caption{Topological diversity of the flows}
    \label{fig:topological-diversity}
\end{figure}

\begin{figure}[h]
    \centering
    \includegraphics[width=1\textwidth]{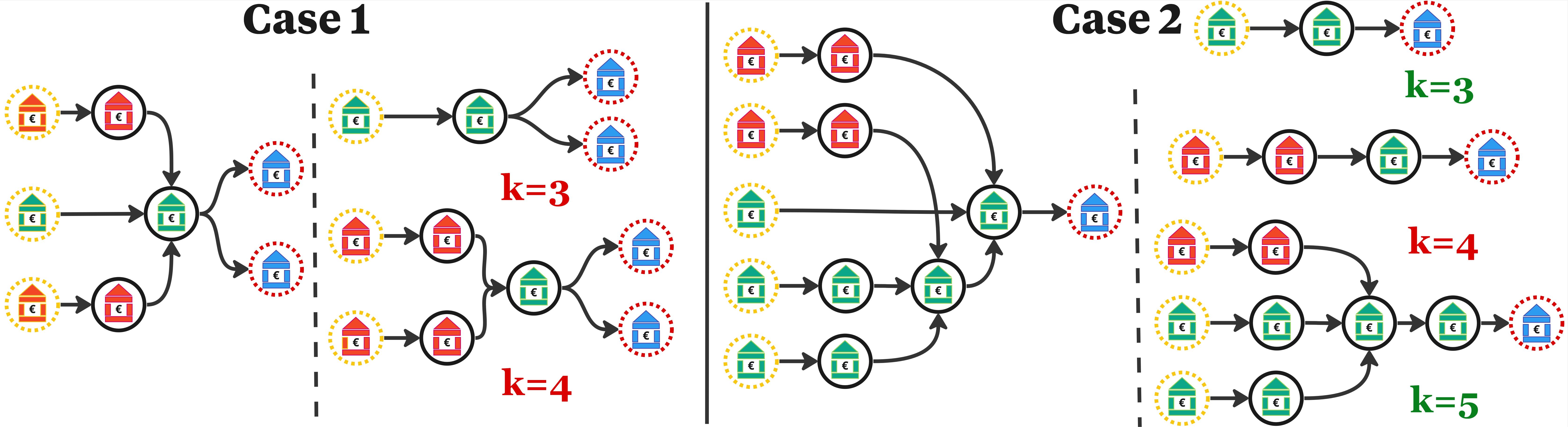}
    \caption{Two cases of \textbf{real} flows. On the left of the dashed lines are the flows detected by {\fastman}, and on the right, the series of \textbf{separate} flows detected by FlowScope. The red font for k=\textit{x} indicates that the flow was not flagged suspicious by FlowScope, based on risk criterion \textit{C3}.}
    \label{fig:cases}
    \vspace{-0.5cm}
\end{figure}
\vspace{-0.1cm}
\subsubsection{Topological Diversity}
Fig. \ref{fig:topological-diversity} shows the (topological) parameter-free nature of {\fastman}. This validates \textit{Motivation 2 and 3} described in Section \ref{sec:related} - i.e., complex money laundering networks involve transactions among several parties, covering longer distances (more than 2 hops). In the left most graph, we are using diameter on the x-axis instead of (number of) hops. This is because {\fastman} can detect flows with varying path lengths. For both the detected (in scope) and the suspicious (cases) flows, we observe a diverse range of values for diameter; and number of dispense and sink accounts. Fig. \ref{fig:cases} shows two complex \textbf{real} flows which will go unnoticed by FlowScope, even after multiple runs with different configurations. FlowScope has a parameter \textit{k} (the \textit{k} in the $k$-partite graph) for controlling the number of hops. When \textit{k=3}, you have 2 hops in the flows; with \textit{k=4}, 3 hops; and so on. For the two cases, {\fastman} detected the complete flows with all the relevant bank accounts involved in contributing towards the suspiciousness of the flows. To detect the same flows using FlowScope, it has to be run with 2 configurations of \textit{k} for the first case; and 3 for the second case. For case 1, both flows will go undetected because individually neither of them qualify for \textit{C3} - i.e., the maximum flow of funds do not reach the minimum threshold set in the criterion. For case 2, the flow detected by FlowScope with \textit{k=4} will go undetected; the other two would qualify. This still makes the quality of the cases lower, as 2 additional suspicious accounts in that flow have gone undetected. We can not further show or (even) summarize all the cases because of \textit{Reason 2} (mentioned in Section \ref{sec:experiments}).

\section{Conclusion and Future Work}
\label{sec:conclusion}
We showed that with {\fastman} it is possible to follow \textit{all suspicious} trails of money, for all nodes. Following Occam's razor, we are making the fewest AML related assumptions as possible. We are putting every conceivable money trail, within a reasonable time window \( \Delta w \), under consideration to be detected as suspicious. Extensive experimental evaluation with two state-of-the-art AML approaches have proved the superiority of {\fastman} in terms of scalability and effectiveness. There are numerous typologies that we can model using {\fastman}. We intend to design a purely community detection -centric model for constructing communities of bank accounts that have recurring flows, in different time periods. Additionally, we plan to apply {\fastman} within the context of the \textit{targeted network search} problem - i.e., use {\fastman} to return all the dominant flows a query account is involved in. Finally, as mentioned in Section \ref{sec:temporal-second}, we plan to do an extensive study for functionally evaluating {\fastman} with higher (than \(2^{nd}\)) and multi-order graph representations.
\clearpage

\printbibliography 

@article{flowscope, title={FlowScope: Spotting Money Laundering Based on Graphs}, volume={34}, url={https://ojs.aaai.org/index.php/AAAI/article/view/5906}, DOI={10.1609/aaai.v34i04.5906}, abstractNote={&lt;p&gt;Given a graph of the money transfers between accounts of a bank, how can we detect money laundering? Money laundering refers to criminals using the bank’s services to move massive amounts of illegal money to untraceable destination accounts, in order to inject their illegal money into the legitimate financial system. Existing graph fraud detection approaches focus on dense subgraph detection, without considering the fact that money laundering involves high-volume &lt;em&gt;flows&lt;/em&gt; of funds through chains of bank accounts, thereby decreasing their detection accuracy. Instead, we propose to model the transactions using a multipartite graph, and detect the complete flow of money from source to destination using a scalable algorithm, FlowScope. Theoretical analysis shows that FlowScope provides guarantees in terms of the amount of money that fraudsters can transfer without being detected. FlowScope outperforms state-of-the-art baselines in accurately detecting the accounts involved in money laundering, in both injected and real-world data settings.&lt;/p&gt;}, number={04}, journal={Proceedings of the AAAI Conference on Artificial Intelligence}, author={Li, Xiangfeng and Liu, Shenghua and Li, Zifeng and Han, Xiaotian and Shi, Chuan and Hooi, Bryan and Huang, He and Cheng, Xueqi}, year={2020}, month={04}, pages={4731-4738} }

@InProceedings{dbj,
author="Starnini, Michele
and Tsourakakis, Charalampos E.
and Zamanipour, Maryam
and Panisson, Andr{\'e}
and Allasia, Walter
and Fornasiero, Marco
and Puma, Laura Li
and Ricci, Valeria
and Ronchiadin, Silvia
and Ugrinoska, Angela
and Varetto, Marco
and Moncalvo, Dario",
editor="Dong, Yuxiao
and Kourtellis, Nicolas
and Hammer, Barbara
and Lozano, Jose A.",
title="Smurf-Based Anti-money Laundering in Time-Evolving Transaction Networks",
booktitle="Machine Learning and Knowledge Discovery in Databases. Applied Data Science Track",
year="2021",
publisher="Springer International Publishing",
address="Cham",
pages="171--186",
isbn="978-3-030-86514-6"
}

@article{smurfing,
 URL = {https://uknowledge.uky.edu/law_facpub/344/},
 author = {Welling, S.N.},
 journal = {Fla. Law},
 pages = {287--343},
 title = {Smurfs, money laundering and the federal criminal law: The crime of structuring transactions},
 volume = {41},
 year = {1989}
}

@online{fatf,
    author = "Financial Action Task Force (FATF)",
    title = "FATF Black and Grey Lists",
    url  = "https://www.fatf-gafi.org/en/countries/black-and-grey-lists.html",
    addendum = "(accessed: 25.03.2023)"
}

@online{acpr,
    author = "Actu IA",
    title = "The ACPR launches an experiment on data sharing to combat money laundering and terrorist financing",
    url  = "https://www.actuia.com/english/the-acpr-launches-an-experiment-on-data-sharing-to-combat-money-laundering-and-terrorist-financing/",
    addendum = "(accessed: 14.06.2023)"
}

@online{tmnl,
    author = "NVB",
    title = "Transaction Monitoring Netherlands: a unique step in the fight against money laundering and the financing of terrorism",
    url  = "https://www.nvb.nl/english/transaction-monitoring-netherlands-a-unique-step-in-the-fight-against-money-laundering-and-the-financing-of-terrorism/",
    addendum = "(accessed: 14.06.2023)",
    sortname = "1"
}

@online{cosmic,
    author = "BIIA",
    title = "Singapore Banks To Share Information Voluntarily To Fight Money Laundering",
    url  = "https://www.biia.com/singapore-banks-to-share-information-voluntarily-to-fight-money-laundering/",
    addendum = "(accessed: 14.06.2023)"
}

@online{aurora,
    author = "BIS",
    title = "BIS concludes Project Aurora, a proof of concept based on the use of data, technology and collaboration to combat money laundering across institutions and borders",
    url  = "https://www.bis.org/about/bisih/topics/fmis/aurora.htm",
    addendum = "(accessed: 14.06.2023)"
}

@online{mule,
    author = "Europol",
    title = "Money Muling",
    url  = "https://www.europol.europa.eu/operations-services-and-innovation/public-awareness-and-prevention-guides/money-muling",
    addendum = "(accessed: 14.06.2023)"
}

@online{mloverview,
    author = "The United Nations Office on Drugs and Crime (UNODC)",
    title = "Money Laundering Overview",
    url  = "https://www.unodc.org/unodc/en/money-laundering/overview.html",
    addendum = "(accessed: 25.03.2023)"
}

@online{iban,
    author = "iban.org",
    title = "International Bank Account Number",
    url  = "https://www.iban.org/",
    addendum = "(accessed: 25.03.2023)"
}

@online{ubo,
    author = "Society for Worldwide Interbank Financial Telecommunications (SWIFT)",
    title = "What is an Ultimate Beneficial Owner",
    url  = "https://www.swift.com/your-needs/financial-crime-cyber-security/know-your-customer-kyc/ultimate-beneficial-owner-ubo",
    addendum = "(accessed: 25.03.2023)"
}

@article{study,
author = {Zhao, Zhongying and Zheng, Shaoqiang and Li, Chao and Sun, Jinqing and Chang, Liang and Chiclana, Francisco},
year = {2018},
month = {06},
pages = {1-10},
title = {A comparative study on community detection methods in complex networks},
volume = {35},
journal = {Journal of Intelligent \& Fuzzy Systems},
doi = {10.3233/JIFS-17682}
}

@article{leiden,
author={Traag, V. A. and Waltman, L. and van Eck, N. J.},
title={From Louvain to Leiden: guaranteeing well-connected communities},
journal={Scientific Reports},
year={2019},
month={03},
day={26},
volume={9},
number={1},
pages={5233},
issn={2045-2322},
doi={10.1038/s41598-019-41695-z},
url={https://doi.org/10.1038/s41598-019-41695-z}
}

@article{Blondel_2008,
	doi = {10.1088/1742-5468/2008/10/p10008},
	url = {https://doi.org/10.1088/1742-5468/2008/10/p10008},
	year = 2008,
	month = {10},
	publisher = {{IOP} Publishing},
	volume = {2008},
	number = {10},
	pages = {P10008},
	author = {Vincent D Blondel and Jean-Loup Guillaume and Renaud Lambiotte and Etienne Lefebvre},
	title = {Fast unfolding of communities in large networks},
	journal = {Journal of Statistical Mechanics: Theory and Experiment}
}

@misc{wagenseller2017size,
  author={Wagenseller, Paul and Wang, Feng and Wu, Weili},
  journal={IEEE Transactions on Computational Social Systems}, 
  title={Size Matters: A Comparative Analysis of Community Detection Algorithms}, 
  year={2018},
  volume={5},
  number={4},
  pages={951-960},
  doi={10.1109/TCSS.2018.2875626}
}

@article{dunbar1993, title={Coevolution of neocortical size, group size and language in humans}, volume={16}, DOI={10.1017/S0140525X00032325}, number={4}, journal={Behavioral and Brain Sciences}, publisher={Cambridge University Press}, author={Dunbar, R. I. M.}, year={1993}, pages={681–694}}

@misc{akoglu2014graphbased,
author={Akoglu, Leman and Tong, Hanghang and Koutra, Danai},
title={Graph based anomaly detection and description: a survey},
journal={Data Mining and Knowledge Discovery},
year={2015},
month={05},
day={01},
volume={29},
number={3},
pages={626-688},
issn={1573-756X},
doi={10.1007/s10618-014-0365-y},
url={https://doi.org/10.1007/s10618-014-0365-y}
}

@article{supervised,
author = {Savage, David and Wang, Qingmai and Chou, Pauline and Zhang, Xiuzhen and Yu, Xinghuo},
year = {2016},
month = {08},
pages = {},
title = {Detection of money laundering groups using supervised learning in networks}
}

@InProceedings{oddball,
author="Akoglu, Leman
and McGlohon, Mary
and Faloutsos, Christos",
editor="Zaki, Mohammed J.
and Yu, Jeffrey Xu
and Ravindran, B.
and Pudi, Vikram",
title="oddball: Spotting Anomalies in Weighted Graphs",
booktitle="Advances in Knowledge Discovery and Data Mining",
year="2010",
publisher="Springer Berlin Heidelberg",
address="Berlin, Heidelberg",
pages="410--421",
isbn="978-3-642-13672-6"
}

@ARTICLE{ego2,
  author={Dumitrescu, Bogdan and Băltoiu, Andra and Budulan, Ştefania},
  journal={IEEE Access}, 
  title={Anomaly Detection in Graphs of Bank Transactions for Anti Money Laundering Applications}, 
  year={2022},
  volume={10},
  number={},
  pages={47699-47714},
  doi={10.1109/ACCESS.2022.3170467}}

@article{SHARMA201744,
title = {Community Detection Algorithm for Big Social Networks Using Hybrid Architecture},
journal = {Big Data Research},
volume = {10},
pages = {44-52},
year = {2017},
issn = {2214-5796},
doi = {https://doi.org/10.1016/j.bdr.2017.10.003},
url = {https://www.sciencedirect.com/science/article/pii/S2214579616302349},
author = {Rahil Sharma and Suely Oliveira}
}

@InProceedings{graph_partitioning,
author="Carlini, Emanuele
and Dazzi, Patrizio
and Esposito, Andrea
and Lulli, Alessandro
and Ricci, Laura",
editor="Lopes, Lu{\'i}s
and {\v{Z}}ilinskas, Julius
and Costan, Alexandru
and Cascella, Roberto G.
and Kecskemeti, Gabor
and Jeannot, Emmanuel
and Cannataro, Mario
and Ricci, Laura
and Benkner, Siegfried
and Petit, Salvador
and Scarano, Vittorio
and Gracia, Jos{\'e}
and Hunold, Sascha
and Scott, Stephen L.
and Lankes, Stefan
and Lengauer, Christian
and Carretero, Jesus
and Breitbart, Jens
and Alexander, Michael",
title="Balanced Graph Partitioning with Apache Spark",
booktitle="Euro-Par 2014: Parallel Processing Workshops",
year="2014",
publisher="Springer International Publishing",
address="Cham",
pages="129--140",
isbn="978-3-319-14325-5"
}

@article{temporal,
author = {Wang, Yishu and Yuan, Ye and Ma, Yuliang and Wang, Guoren},
year = {2019},
month = {12},
pages = {1-15},
title = {Time-Dependent Graphs: Definitions, Applications, and Algorithms},
volume = {4},
journal = {Data Science and Engineering},
doi = {10.1007/s41019-019-00105-0}
}

@article{geometry,
author = {Granados, Oscar and Vargas, Andrés},
year = {2022},
month = {12},
pages = {},
title = {The geometry of suspicious money laundering activities in financial networks},
volume = {11},
journal = {EPJ Data Science},
doi = {10.1140/epjds/s13688-022-00318-w}
}

@inproceedings{Scholtes_2017,
	doi = {10.1145/3097983.3098145},
	url = {https://doi.org/10.1145/3097983.3098145},
	year = 2017,
	month = {08},
	publisher = {{ACM}
},
	author = {Ingo Scholtes},
	title = {When is a Network a Network?},
	booktitle = {Proceedings of the 23rd {ACM} {SIGKDD} International Conference on Knowledge Discovery and Data Mining}
}

@inproceedings{graphframes,
author = {Dave, Ankur and Jindal, Alekh and Li, Li and Xin, Reynold and Gonzalez, Joseph and Zaharia, Matei},
year = {2016},
month = {06},
pages = {1-8},
title = {GraphFrames: an integrated API for mixing graph and relational queries},
doi = {10.1145/2960414.2960416}
}

@inproceedings{spark,
author = {Zaharia, Matei and Chowdhury, Mosharaf and Das, Tathagata and Dave, Ankur and Ma, Justin and McCauley, Murphy and Franklin, Michael J. and Shenker, Scott and Stoica, Ion},
title = {Resilient Distributed Datasets: A Fault-Tolerant Abstraction for in-Memory Cluster Computing},
year = {2012},
publisher = {USENIX Association},
address = {USA},
booktitle = {Proceedings of the 9th USENIX Conference on Networked Systems Design and Implementation},
pages = {2},
numpages = {1},
location = {San Jose, CA},
series = {NSDI'12}
}

@article{multiplex,
  title = {Mathematical Formulation of Multilayer Networks},
  author = {De Domenico, Manlio and Sol\'e-Ribalta, Albert and Cozzo, Emanuele and Kivel\"a, Mikko and Moreno, Yamir and Porter, Mason A. and G\'omez, Sergio and Arenas, Alex},
  journal = {Phys. Rev. X},
  volume = {3},
  issue = {4},
  pages = {041022},
  numpages = {15},
  year = {2013},
  month = {12},
  publisher = {American Physical Society},
  doi = {10.1103/PhysRevX.3.041022},
  url = {https://link.aps.org/doi/10.1103/PhysRevX.3.041022}
}

@article{modularity,
author = {M. E. J. Newman },
title = {Modularity and community structure in networks},
journal = {Proceedings of the National Academy of Sciences},
volume = {103},
number = {23},
pages = {8577-8582},
year = {2006},
doi = {10.1073/pnas.0601602103},
URL = {https://www.pnas.org/doi/abs/10.1073/pnas.0601602103},
eprint = {https://www.pnas.org/doi/pdf/10.1073/pnas.0601602103}}

@article{10.14778/3229863.3229874,
author = {Qiu, Xiafei and Cen, Wubin and Qian, Zhengping and Peng, You and Zhang, Ying and Lin, Xuemin and Zhou, Jingren},
title = {Real-Time Constrained Cycle Detection in Large Dynamic Graphs},
year = {2018},
issue_date = {August 2018},
publisher = {VLDB Endowment},
volume = {11},
number = {12},
issn = {2150-8097},
url = {https://doi.org/10.14778/3229863.3229874},
doi = {10.14778/3229863.3229874},
journal = {Proc. VLDB Endow.},
month = {08},
pages = {1876–1888},
numpages = {13}
}

@misc{peng2022tdb,
  author={Peng, You and Lin, Xuemin and Yu, Michael and Zhang, Wenjie and Qin, Lu},
  booktitle={2023 IEEE 39th International Conference on Data Engineering (ICDE)}, 
  title={TDB: Breaking All Hop-Constrained Cycles in Billion-Scale Directed Graphs}, 
  year={2023},
  volume={},
  number={},
  pages={137-150},
  doi={10.1109/ICDE55515.2023.00018}
}

@ARTICLE{9762926,
  author={Dumitrescu, Bogdan and Băltoiu, Andra and Budulan, Ştefania},
  journal={IEEE Access}, 
  title={Anomaly Detection in Graphs of Bank Transactions for Anti Money Laundering Applications}, 
  year={2022},
  volume={10},
  number={},
  pages={47699-47714},
  doi={10.1109/ACCESS.2022.3170467}}

@misc{elliott2019anomaly,
      title={Anomaly Detection in Networks with Application to Financial Transaction Networks}, 
      author={Andrew Elliott and Mihai Cucuringu and Milton Martinez Luaces and Paul Reidy and Gesine Reinert},
      year={2019},
      eprint={1901.00402},
      archivePrefix={arXiv},
      primaryClass={stat.AP}
}

@inproceedings{Paranjape_2017,
	doi = {10.1145/3018661.3018731},
	url = {https://doi.org/10.1145%2F3018661.3018731},
	year = 2017,
	month = {02},
	publisher = {{ACM}},
	author = {Ashwin Paranjape and Austin R. Benson and Jure Leskovec},
	title = {Motifs in Temporal Networks},
	booktitle = {Proceedings of the Tenth {ACM} International Conference on Web Search and Data Mining}
}

@inproceedings{spanning,
author = {Huang, Silu and Fu, Ada and Liu, Ruifeng},
year = {2015},
month = {05},
pages = {419-430},
title = {Minimum Spanning Trees in Temporal Graphs},
doi = {10.1145/2723372.2723717}
}

@article{10.14778/2732939.2732945,
author = {Wu, Huanhuan and Cheng, James and Huang, Silu and Ke, Yiping and Lu, Yi and Xu, Yanyan},
title = {Path Problems in Temporal Graphs},
year = {2014},
issue_date = {May 2014},
publisher = {VLDB Endowment},
volume = {7},
number = {9},
issn = {2150-8097},
url = {https://doi.org/10.14778/2732939.2732945},
doi = {10.14778/2732939.2732945},
journal = {Proc. VLDB Endow.},
month = {05},
pages = {721–732},
numpages = {12}
}

@article{temporal-shortest,
author = {Wu, Huanhuan and Cheng, James and Ke, Yiping and Huang, Silu and Huang, Yuzhen and Hejun, Wu},
year = {2016},
month = {11},
pages = {1-1},
title = {Efficient Algorithms for Temporal Path Computation},
volume = {28},
journal = {IEEE Transactions on Knowledge and Data Engineering},
doi = {10.1109/TKDE.2016.2594065}
}

@online{unixts,
    author = "UnixTime.org",
    title = "Unix Timestamp",
    url  = "https://unixtime.org/",
    addendum = "(accessed: 21.06.2023)"
}

@ARTICLE{glad,
  author={Ma, Xiaoxiao and Wu, Jia and Xue, Shan and Yang, Jian and Zhou, Chuan and Sheng, Quan Z. and Xiong, Hui and Akoglu, Leman},
  journal={IEEE Transactions on Knowledge and Data Engineering}, 
  title={A Comprehensive Survey on Graph Anomaly Detection with Deep Learning}, 
  year={2021},
  volume={},
  number={},
  pages={1-1},
  doi={10.1109/TKDE.2021.3118815}}

@online{ubo-data,
    author = "Moody's",
    title = "UBOs: what they are, disclosure requirements, and the data challenge",
    url  = "https://www.moodys.com/web/en/us/kyc/resources/insights/ubos-what-they-are-disclosure-requirements-data-challenge.html",
    addendum = "(accessed: 22.06.2023)",
    sortname = "ubo-moodys"
}

@article{10.1145/1541880.1541882,
author = {Chandola, Varun and Banerjee, Arindam and Kumar, Vipin},
title = {Anomaly Detection: A Survey},
year = {2009},
issue_date = {July 2009},
publisher = {Association for Computing Machinery},
address = {New York, NY, USA},
volume = {41},
number = {3},
issn = {0360-0300},
url = {https://doi.org/10.1145/1541880.1541882},
doi = {10.1145/1541880.1541882},
journal = {ACM Comput. Surv.},
month = {07},
articleno = {15},
numpages = {58}
}

@INPROCEEDINGS{5496972,
  author={Shvachko, Konstantin and Kuang, Hairong and Radia, Sanjay and Chansler, Robert},
  booktitle={2010 IEEE 26th Symposium on Mass Storage Systems and Technologies (MSST)}, 
  title={The Hadoop Distributed File System}, 
  year={2010},
  volume={},
  number={},
  pages={1-10},
  doi={10.1109/MSST.2010.5496972}}

\end{document}